\documentclass[letterpaper, 10 pt, conference]{ieeeconf}  

\IEEEoverridecommandlockouts                              

\usepackage{amsmath} 
\usepackage{amssymb}  
\usepackage{graphicx} 
\usepackage{todonotes}
\usepackage{booktabs}
\usepackage{multirow}
\usepackage{float}
\usepackage{balance}
\usepackage{censor}
\usepackage{hyperref}
\usepackage{cite}

\title{\LARGE \bf
Distilling Global Traversability Priors for Image-based Affordance Prediction in Off-road Environments
}

\author{Matthew Sivaprakasam$^{1}$, Samuel Triest$^{1}$, Micah Nye$^{1}$, Deegan Atha$^{2}$,\\ Shehryar Khattak$^{2}$, David Fan$^{2}$, Wenshan Wang$^{1}$, and Sebastian Scherer$^{1}$ 
\thanks{$^{1}$ Robotics Institute, Carnegie Mellon University}
\thanks{msivapra,striest,micahn,wenshanw,basti@andrew.cmu.edu }%
\thanks{$^{2}$ Field AI}}

\begin{document}

\maketitle
\thispagestyle{empty}
\pagestyle{empty}

\begin{figure*}[]
	\centering
	\includegraphics[width=.95\linewidth]{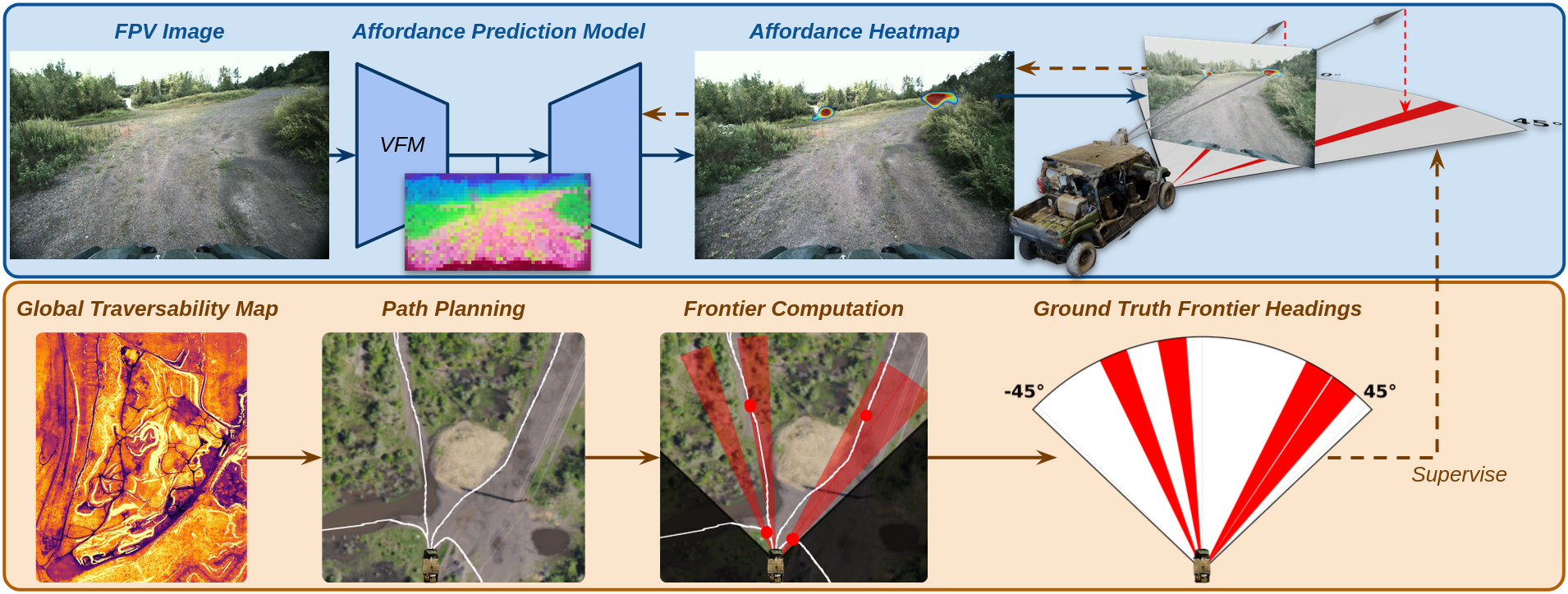}
	\caption{
      Overview of our training pipeline. FPV images are fed into a VFM, the output of which in turn is fed into a decoder that produces an affordance heatmap. This heatmap is projected into frontier headings using camera intrinsics. These are then supervised using ground truth frontiers computed based on a satellite traversability map aligned with the robot's own experience.
     }
	\label{fig:overview}
    \vspace{-.5cm}
\end{figure*}

\begin{abstract}


Standard methods for autonomous navigation in unstructured terrain are prone to myopic behaviors in long-horizon scenarios. The use of metric maps built from LiDAR or cameras provides necessary local geometry and semantic information but is strictly limited by depth sensing range. By discarding data beyond the mapping horizon robots suffer from suboptimal, short-sighted decisions. To recover this lost information, we focus on extracting long-range traversability-aware frontiers directly from first-person-view (FPV) images. By leveraging satellite imagery, we compute the set of feasible navigation paths for a dataset of image/pose pairs and use them to supervise our network, reducing the need for extensive human demonstration data. We demonstrate that this approach improves performance in long-range off-road navigation over existing methods by more than 10\% in various offline benchmarks and reduces the number of human interventions incurred in a set of real-world experiments. More details can be found at \href{https://theairlab.org/ss_frontiers_iros}{https://theairlab.org/ss\_frontiers\_iros}.
\end{abstract}


\section{Introduction}
Long-range information is critical for robust and efficient robot performance in off-road environments. Ideally this comes from GPS and global map data, but in practice this information is not available at deployment. Instead, the robot must make local approximations using its on-board sensors in order to navigate successfully. This is often done by first aggregating sensor data from camera and/or LiDAR into a local map or representation (often metric and Birds-Eye-View (BEV)), on which methods such as obstacle detection and traversability analysis are performed. The robot then determines a path that is locally feasible while still making progress towards the goal. 

This has been shown to be effective for short-horizon problems. If the goal is sufficiently close, all relevant planning information is contained in the local map. When the goal is further, however, the long-range information from the sensors necessary for navigation is often discarded. This challenge stems from the choice of using BEV maps alone as the planning representation, as metric maps struggle with computational tractability and sensor noise with increased range. Several works attempt to circumvent this by learning to predict plans directly from FPV images \cite{sridhar2023nomad}. Although this works with simpler geometry and terrain, its success has been limited in off-road scenarios \cite{thakker2025riskguided, schmittle2025long, guamancastro2025vamos}. More recent methods have shown promise in leveraging both FPV and BEV representations together \cite{schmittle2025long, thakker2025riskguided, Deva2020Learning, kim2025raven}. For example, relevant long-range cues can first be extracted from FPV information before it is projected into the BEV. The local planner is then able to obey the long-range cues to avoid myopic behaviors while still using the local BEV map to explicitly reason about 3D geometry. These long-range cues are trained in a hindsight manner that often relies on depth to project robot future trajectories or demonstrations into the FPV image, but state-of-the-art depth estimation methods still struggle in extreme off-road scenarios. Long Range Navigator (LRN) \cite{schmittle2025long} eliminates this depth requirement by instead using off-the-shelf point tracking models, but can struggle with noisy or failed point tracks. Moreover, relying solely on human demonstrations fails to efficiently capture the multi-modal nature of navigation. For example, in scenarios where the robot encounters a "fork in the road," where two paths are viable, the demonstration data may only show one valid path, leaving the robot uncertain about the other.

To address these challenges, we propose a new method for long-range affordance learning and frontier estimation. We generate supervision data by first generating continuous-valued traversability maps through an extension of SALON \cite{salon}. From these maps we generate several feasible long-range paths per pose, providing a more thorough set of demonstrations than human data alone. Representing these paths in polar coordinates allows us to supervise in the heading-space, circumventing the issues of depth-estimation noise and failed point tracks that hinder existing image-space methods. Through this method we present four key contributions:

\begin{enumerate}
\item An improved approach for learning long-range affordances that is traversability-aware and optimizes the final heading-space representation directly, without requiring expensive hand-labels.
\item An extension of SALON that allows for robot-specific global traversability maps from satellite imagery.
\item Supplementation of demonstration data with planning-based labels to improve data efficiency for off-road navigation tasks.
\item A suite of real-world and offline experiments that demonstrate viability of our approach in different environments and sensor configurations.
\end{enumerate}




\section{Related Work}

\subsection{Long-Range Off-Road Autonomy}
Many recent state-of-the-art approaches to off-road autonomy focus on estimating traversability from local maps with supervision from hand-crafted cost functions \cite{meng2023terrainnet, step}, human demonstration \cite{triest2024velociraptor, zhang2025crestescalablemaplessnavigation}, or proprioceptive feedback \cite{salon, mattamala25wild, hdif}. While robust local traversability estimation is crucial, these works do not properly account for information beyond the local map.

Long-range reasoning is a well-researched problem in robotics, but often under assumptions that do not hold in off-road environments. Some works reason about sub-goals at the edge of local perception range, but assume access to a map of the entire environment \cite{valueofplanning-hatch} or assume a simplified environment representation such as a 2D occupancy grid \cite{harutyunyan2025mapexrl, jeric2025dare,pmlr-v87-stein18a}. Other methods do use images as input but either require depth \cite{Deva2020Learning} or only predict paths with a limited range \cite{sridhar2023nomad, hirose2025omnivla, zhang2026ventura}. 

Most recently, success has been found in approaches that involve both long-range reasoning and local map representations \cite{kim2025raven}. For example, traversability can be reasoned about in the image space and then converted to a BEV map of radial bins using either explicit \cite{hadsell-learning} or learned projections \cite{fahnestock}. While learning the projections improves accuracy at range, it is less likely to generalize across camera configurations. Other works extend on existing image-based planning models by guiding the model with local traversability information, but are still limited by the range of the planning model itself \cite{thakker2025riskguided}. Most relevant to our work is Long Range Navigator \cite{schmittle2025long}, which uses CoTracker \cite{karaev2023cotracker} to automatically label images and explicitly accounts for camera intrinsics. However, this method relies on sufficient demonstration data and lack of noise in the CoTracker outputs.


\subsection{Traversability from Satellite Imagery}
When satellite imagery is available, generating global traversability maps that a robot can then plan on and relate to using GPS is a popular approach. One way to accomplish this is with a human in the loop, either providing expert demonstration data \cite{Ratliff-2009-10259} or specifying user preferences at run-time \cite{mao_PACER_RAL2025}. However, relying heavily on human labeling reduces scalability. Other approaches circumvent human labeling by using proprioceptive feedback from the robot \cite{eder, 11128264} but either rely on discrete semantic classes or operate at a low resolution. All of these approaches also rely on access to the satellite map during deployment. Methods do exist that train models to reconstruct satellite-imagery-derived information from local sensor information but they either don't apply this to traversability estimation \cite{ho2024map} or train on incomplete traversability maps \cite{mm-roadrunner}.


\section{Problem Setup}

We consider the problem of a robot navigating from a start position $x_1$ to goal $x_G$. In order to do so it computes a path $\tau$ minimizing the expected cost $C$ incurred from traveling between the two points:

\begin{align}
    J(\tau) &= \mathop{\mathbb{E}}\left[ \sum_{t=1}^{G} C(x_t) \right]
    \label{ref:eq_globalJ}
\end{align}

In reality, the robot does not have access to the true cost function $C$, and must instead infer it from local sensor observations $O_t$ using a heuristic function or learned model $C_\theta$. Due to computational constraints and limited sensing range, it is impractical to try to compute the whole path $\tau$ directly. Instead, a local policy $\pi$ is run at each timestep to find a shorter-range path $\hat{\tau}$ to a fixed horizon $H<<G$ via Model Predictive Control (MPC). However, this approach to navigation is myopic as it does not sufficiently reason about the whole environment. Prior work addresses this by incorporating global information into the terminal cost \cite{valueofplanning-hatch}, such that the policy now minimizes:

\begin{align}
   \hat{J}(\hat{\tau}) &=
    \mathop{\mathbb{E}}
    \left[
    \sum_{t=1}^{H} C_\theta(x_t | O_t)
    + V(x_H)
    \right]
    \label{ref:eq_localJ}\\
    \pi^* &= \arg\min_{\pi} \hat{J}(\hat{\tau})
    \label{ref:eq_tau*}
\end{align}

where $V$ provides the cost-to-go. We focus on learning an approximation $V_\theta(x_H|O_t)$ and leverage existing work to compute $C_\theta$.
We adopt an approach similar to LRN and first compute $V_k$ for a set of $K$ angular bins with heading $\theta_k$ around the robot. $V_k$ is decomposed as:

\begin{align}
    V_k(x_1, x_G, x_H) &= A_k(x_1, x_H)D_k(x_H,x_G)
    \label{ref:eq_Vk_decompose}\\
    V(x_H) &= V_{k^*}, k^*=\arg\min_{k}d(x_H, \theta_k)
    \label{ref: eq_Vk2V}
\end{align}

where $d$ computes the angular distance between a state and heading, $D$ is a goal-conditioned navigation cost, and $A$ is a learned goal-agnostic function that predicts a traversability-aware affordability score:

\begin{align}
    A_k(x_1, x_H) &= P_k(x_1, x_H)P_k(x_H, x_G)C^*(x_1, x_G) 
    \label{ref:eq_Ak}
\end{align}
where $P_k(x_a,x_b)$ is the probability that an optimal path from $x_a$ to $x_b$ at $H$ lies in the $k$-th bin. $C^*$ computes the traversability cost of the path between two points according to $C$.

\begin{figure}[]
	\centering
	\includegraphics[width=1.\linewidth]{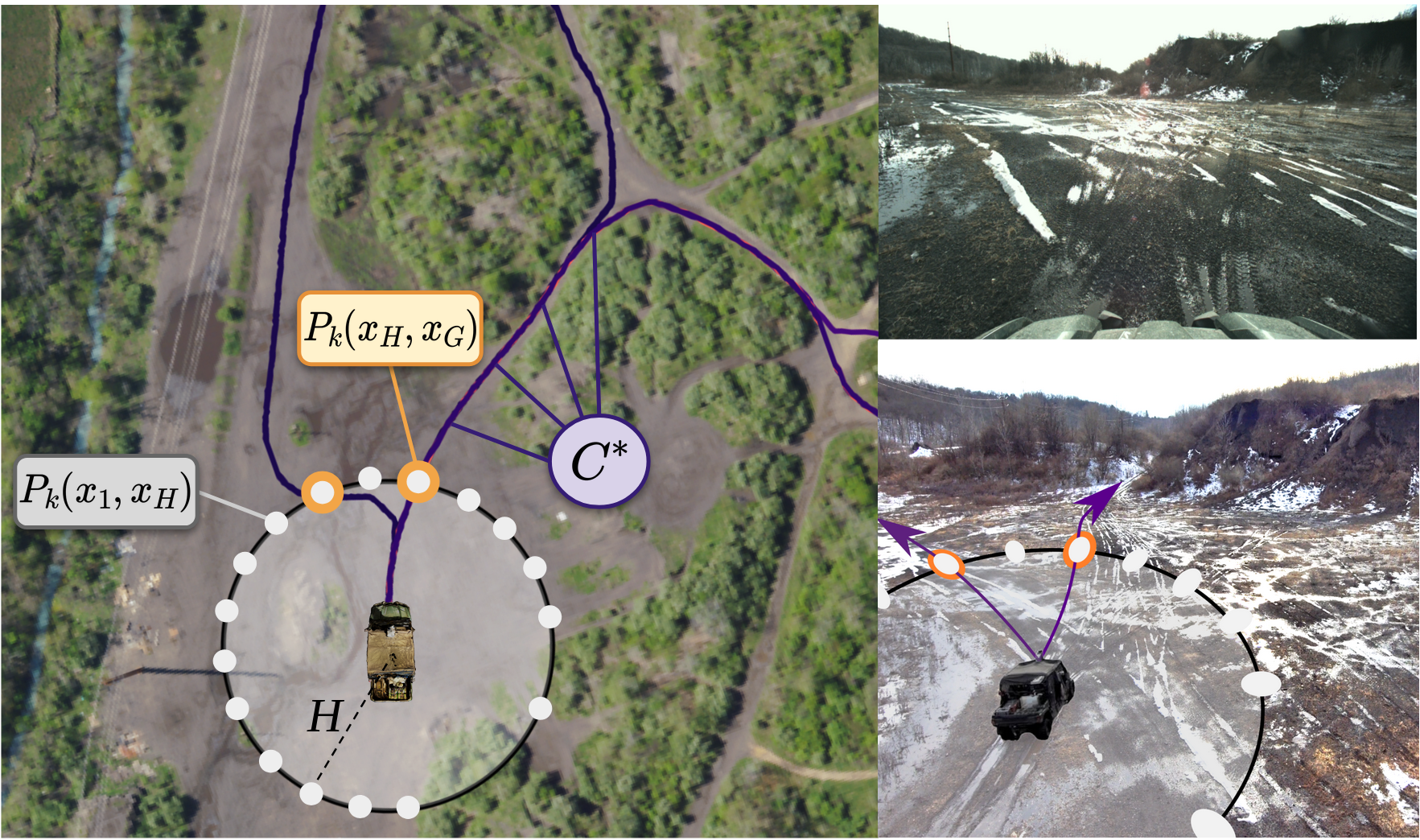}
	\caption{
       Visualization of the components $P_k$ and $C^*$ needed to compute $V_k$ via eq. 4-6. Given a set of optimal long-range paths, $C^*$ models their traversability costs and $P_k$ models its probability of existence. 
     }
	\label{fig:problem_setup}
\end{figure}

\section{Method}

Shown in Fig. \ref{fig:overview}, our process of learning affordability scores $A_{1:K}$ directly from FPV imagery consists of three main components, connected in the heading space. The first stage is the affordance model itself, which uses known camera intrinsics to project affordance heatmaps to heading scores. For training, we convert publicly available satellite imagery to traversability maps. These are used to generate feasible long-range paths, from which traversability-aware frontier headings can be derived for supervision.

\subsection{Affordance Prediction Model}
\label{sec:affordance_prediction_model}
We learn a model that maps sensor observations into an affordance score for each heading bin, using onboard RGB camera images as input (Fig. \ref{fig:overview}). The image $I$ is fed into a visual foundation model (VFM) $f^{VFM}$, which serves as a strong pre-trained feature extractor, the output of which is then fed into a smaller model $f^A_\theta$ trained to predict a traversability-aware affordance image $I_A$:

\begin{align}
    f^A_\theta(f^{VFM}(I_{3\times H \times W})) = I_A \in \mathop{\mathbb{R}^{1 \times H' \times W'}}
    \label{ref:eq_image_model}
\end{align}

We first use the known camera intrinsics to find the corresponding heading for each pixel $p$. $A_k$ (eq. \ref{ref:eq_Ak}) is then estimated by taking a weighted average across all pixels whose heading corresponds to the $k$-th angular bin:


\begin{align}
    A_k = \frac{\sum_{p \in \mathcal{P}_k} e^p \cdot p}{\sum_{p \in \mathcal{P}_k} e^p}, \quad\mathcal{P}_k = \{ p \in I_A \mid \theta_p \approx \theta_k \}
    \label{ref:eq_Pk}
\end{align}


\subsection{Ground Truth Supervision from Satellite Traversability Maps}

Rather than supervise in the image space as in prior methods, we choose to supervise the heading-space representation $A_k$ that is directly used for planning. This is feasible since the process involved in computing $A_k$ from affordance image $I_A$ is fully differentiable, but now requires ground-truth data in heading-space to supervise against. To achieve this, we develop a pipeline that leverages existing off-road driving datasets in conjunction with publicly available satellite imagery to determine which headings contain a frontier.

\begin{figure*}[]
	\centering
	\includegraphics[width=.98\linewidth]{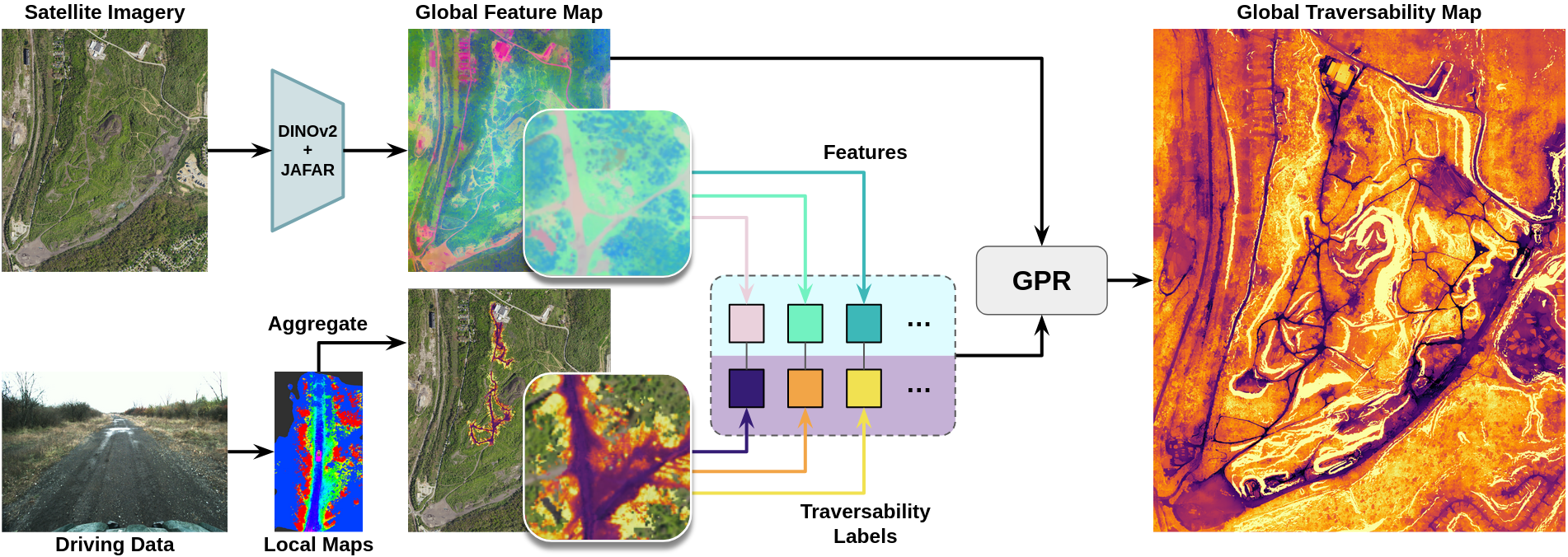}
	\caption{
       Global traversability map computation: We start with an off-road driving dataset, used to generate local traversability maps at each timestep and register them together using GPS. In parallel, high-resolution features are extracted from a satellite map using DINOv2 and JAFAR. Using the registered map as supervision, Gaussian Process Regression (GPR) is used to predict traversability values for the whole map, including areas unvisited by the robot.
     }
	\label{fig:global_map}
    \vspace{-.3cm}
\end{figure*}

\subsubsection{Local Traverability Map Registration}
We first apply SALON to a dataset of real-world vehicle trajectories and corresponding sensor observations. This provides a set of local traversability maps derived from the actual vehicle experience as it drives through challenging terrain. Using RTK GPS data, these maps are then registered into a global map. This global map has a one-to-one mapping with its corresponding RGB satellite map, such that it provides ground truth traversability labels for pixels in the satellite map that are within the local mapping radius of the vehicle's trajectory.

\subsubsection{Global Traversability Map Generation}
In order to generate supervision beyond demonstration data, we require traversability estimates for the \textit{entire} satellite map, not just along the vehicle trajectory. A given satellite map can span thousands of pixels in each direction, so we first divide it into overlapping patches. Visual features are then extracted for each of these RGB patches by feeding it through DINOv2 \cite{oquab2023dinov2} and then upsampled using JAFAR \cite{couairon2025jafar}. These visual feature patches are then aggregated back into a global map using a distance-weighted sliding window.

With visual features extracted for the whole satellite map, we acquire  a set of feature-traversability pairs from regions of the map with known traversability and fit a Gaussian Process (GP) model. This model is then used to regress traversability values for the remaining pixels in the satellite map, effectively propagating the robot's experience across the whole global map. Sharp changes in terrain elevation (such as cliffs) aren't always visible in satellite imagery, so we also use digital elevation models to locate regions with steep slopes and set them as lethal in the global map. We outline this whole process in Fig. \ref{fig:global_map}.


\subsection{Ground Truth Frontier Computation}
Given the global traversability map we can now compute feasible paths, beyond demonstration alone, that the robot can travel. For each timestep in the dataset, we use the vehicle's corresponding GPS location as a starting point from which we apply the Fast Marching Method (FMM) \cite{fastmarch} to compute optimal paths to a diverse set of goal locations. These goals are uniformly distributed along a circle with radius $R>>H$, ensuring that the resulting paths extend well beyond the local planning range.

\begin{figure}[]
	\centering
	\includegraphics[width=.92\linewidth]{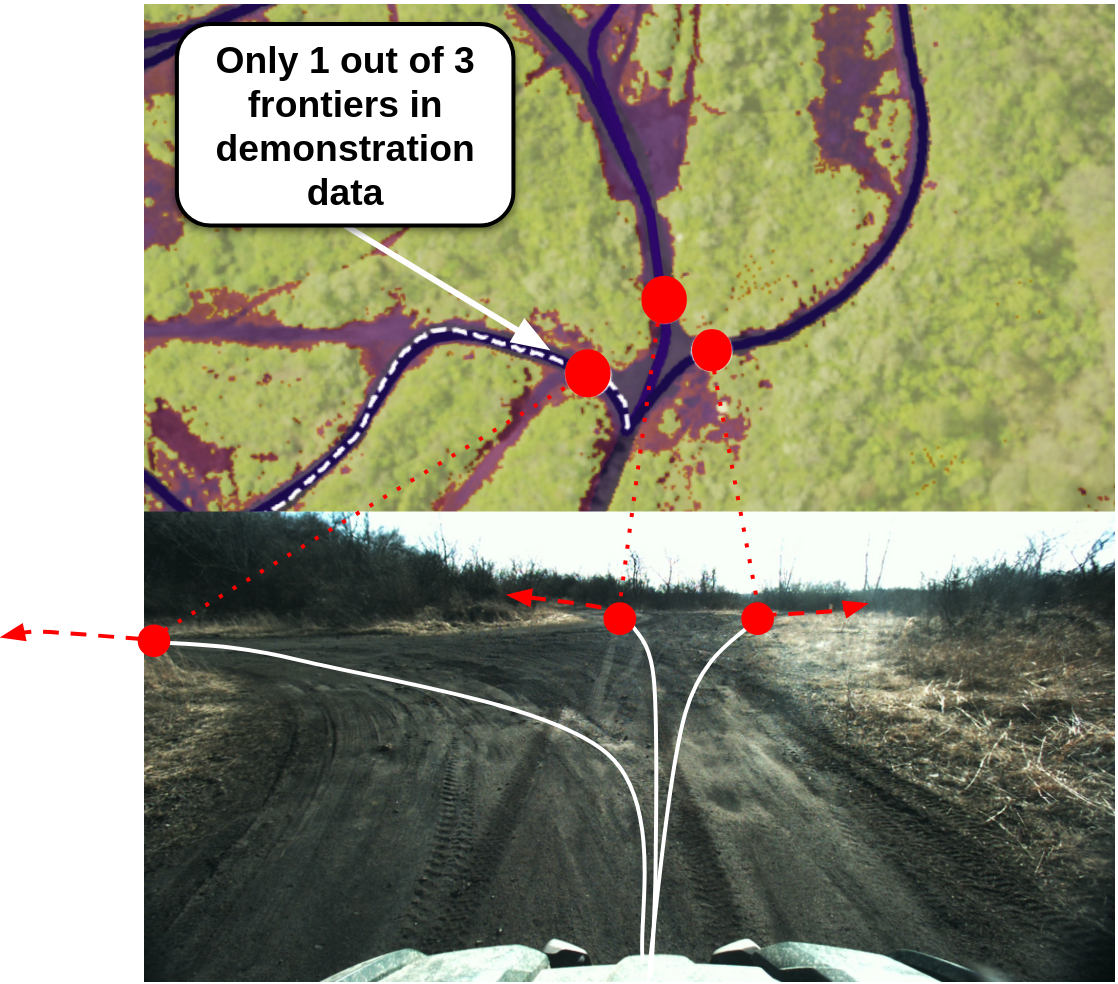}
	\caption{
        Instead of relying solely on demonstration (shown in dashed white), we also plan paths on the global traversability map to a set of long-range goals. For each path, we find the furthest point (red dot) that isn't occluded by lethal terrain and label its relative heading as a viable frontier with a score based its traversability. Corresponding paths are hand-drawn in the FPV image for clarity, with dashed red lines indicating occluded segments.
     }
	\label{fig:gt_gen}
    \vspace{-.5cm}
\end{figure}

For each path generated in this manner we define its frontier location as the furthest non-occluded point along the path, where a point is non-occluded if it lies within the camera's field of view, and no lethal terrain exists along the line between it and the vehicle's starting position (Fig. \ref{fig:gt_gen}). We also compute the traversability of the path by taking the average traversability of each point according to the global map. By converting the frontier location from cartesian to polar coordinates, this process tells us for each heading $\theta_k$ whether or not a viable frontier exists ($P_k$) as well as the expected traversability of the optimal path beyond that frontier ($C^*$), from which $A_k$ can be constructed (eq. \ref{ref:eq_Ak}). In the event that a given heading has multiple viable paths, we optimistically assign it the value from the path with the highest traversability.

Notably, creating supervision this way naturally provides multi-modal synthetic demonstration data. As shown Fig. \ref{fig:gt_gen}, the available expert demonstration data goes along only one path, while our approach of explicitly planning to a broad set of goals allows us to discover alternative paths (or lack thereof).

\subsection{Model \& System Implementation}
\subsubsection{Training}
For our experiments, we use RADIOv3-b \cite{radio} as $f^{VFM}$ to extract features, as it runs in real time on our system and provides semantically rich spatial features distilled from larger VFMs. We keep this frozen and train the lightweight upsampling decoder as $f^\theta_A$. The predicted heatmap $I_A$ produced by $f^A_\theta$ is then reduced to $A_{1:K}$, as described in section \ref{sec:affordance_prediction_model}, and compared to the ground truth $A^*_{1:K}$ Fig. \ref{fig:overview}. We convert both into categorical distributions, and compute the Earth Mover's Distance (EMD) between their probability mass functions $p_A, p_{A^*}$ as the primary loss alongside a small regularization loss on $A_k$:

\begin{align}
    \mathcal{L} = \frac{1}{K^2} \| \textbf{cs}(p_A) - \textbf{cs}(p_{A^*}) \|_2 + \lambda_{reg} \sum_{k=1}^K \left| A_k \right|
    \label{ref:eq_loss}
\end{align}

where \textbf{cs} computes the cumulative sum. We find that using EMD as a loss helps account for the geometric relationship between angular bins \cite{cai2024evora, emdloss}, and the regularization term helps reduce noise in the image.

We train on a dataset consisting of a subset of TartanDrive 2.0 \cite{tartandrive2}, as well as some additional data from a different location, totaling 28,000 samples.

\subsubsection{Deployment}
For real-world deployment, we use SALON \cite{salon}, an open source package that uses a local map of visual features, to predict local traversability maps needed to estimate $C(x_{1:H})$. The visual features from RADIOv3-b are shared between SALON and our affordance prediction module, such that the only overhead added by this method is from $f^A_\theta$. 

$A_k$ is goal-agnostic, so we use $D_k$ to condition $V_k$ on the current goal. We use the same approach as LRN to compute $D_k$ based on a Gaussian centered on the goal heading, such that $D_k$ decreases as $\left|\theta_k - \theta_G \right|$ increases. A goal heading is then chosen based on the max value in $V_{1:K}$ and a waypoint is placed at distance H along the goal heading. Model Predictive Path Integral Control (MPPI) \cite{mppi} is used to travel to this intermediate waypoint while respecting the local traversability maps produced by SALON.

\section{Experiments}

\subsection{Baselines}
We evaluate our method against three baselines:
\begin{enumerate}
    \item \textit{No Frontier Estimation (NFE):} No predicted frontiers are used as MPPI calculates paths. Only the local traversability maps from SALON are used.
    \item \textit{Long Range Navigator (LRN):} We train a model with the same architecture as ours but trained using the approach from Long Range Navigator. We run CoTracker \cite{karaev2023cotracker} on our dataset to automatically generate ground truth affordance heatmaps which are used to supervise the model in the image space.
    \item \textit{Demonstration-only Supervision (DOS)} We train our method only using demonstration data to compute frontiers, rather than generating additional samples using the global planner. Since this formulation doesn't provide labels for frontiers that the demonstration doesn't pass through, we replace the EMD loss with an MSE loss as well as a loss that encourages the demonstrated frontier heading to have a higher score than the other headings. 
\end{enumerate}

\subsection{Offline Evaluation} 

\subsubsection{Image-space Affordance Classification Sub-task}
Even though our method is not supervised in the image space, we still desire that it learns to produce semantically meaningful predictions. To test this, we hand-label a set of 250 images $I_A^*$ following the same process from LRN, and binarize the predicted affordance images $I_A$ in order to compare them to the labeled images.
The results of this analysis are shown in Table \ref{tab:heatmap_metrics}, where "similar" environment consists of images from a held-out test set in the same training environment, and "OOD" refers to an out-of-distribution environment and using a camera with a wider field-of-view (rightmost column in Fig. \ref{fig:img_preds})

\subsubsection{Heading-space Classification and Planning Sub-task }
\label{sec:heading_space_tasks}

Assuming an accurate global traversability map, we aim to evaluate the optimality of paths produced by using our method within a navigation stack. Given a start pose and its corresponding FPV image, we compute a privileged ground-truth path by planning over the full global map to a distant goal (chosen using future demonstration data to guarantee feasibility). The model then predicts a frontier heading from the FPV image, based on its own predictions and relative heading to the goal. To approximate deployment conditions, we restrict planning to a local circular region of the map with radius $H$ centered at the start position. Within this cropped map, we generate a new plan from the start to a point on the boundary aligned with the predicted frontier heading. We quantify performance by comparing this locally planned path against the privileged path computed using the full traversability map. We evaluate for two different types of tasks: classification and goal-conditioned frontier selection. Classification metrics describe model's ability to correctly detect all frontier headings. Goal-conditioned metrics aim to compare the local path to the ground-truth path by computing the heading error between the two at distance $H$, and Modified Hausdorff Distance (MHD) between the local path and the global path up to $H$. 

\subsection{Real-World Evaluation} 
In addition to offline evaluation we also test our methods against the baselines in a series of real-world experiments, where the robot must navigate autonomously to a goal point far away enough that it can't reach it by travelling in a straight line or simple trail following. We create five core scenarios and observe the time to complete the mission and total number of human interventions. A human intervention occurs either when the robot is about to drive over or into lethal terrain or when it commits to a route that has no feasible route to the goal that doesn't involve doubling back. Upon intervention, the operator corrected the vehicle by driving it to a point immediately after the incorrect decision point such that the vehicle could resume autonomy in the correct direction. We measure time to include the time taken for the human operator to correct the vehicle as this reflects time lost during operation in downstream applications.

The direct distances to the goal points range from 160-350 meters away from the start, with actual driven distances ranging from 250-400 meters due to obstructions and trails. In total, the robot must drive over 1500 meters in order to complete all five courses.

The platform used is a Yamaha Viking All-Terrain Vehicle (ATV), retrofitted with actuation hardware and a sensor payload including a LiDAR, camera, and IMU.

\begin{figure*}[]
	\centering
	\includegraphics[width=.99\linewidth]{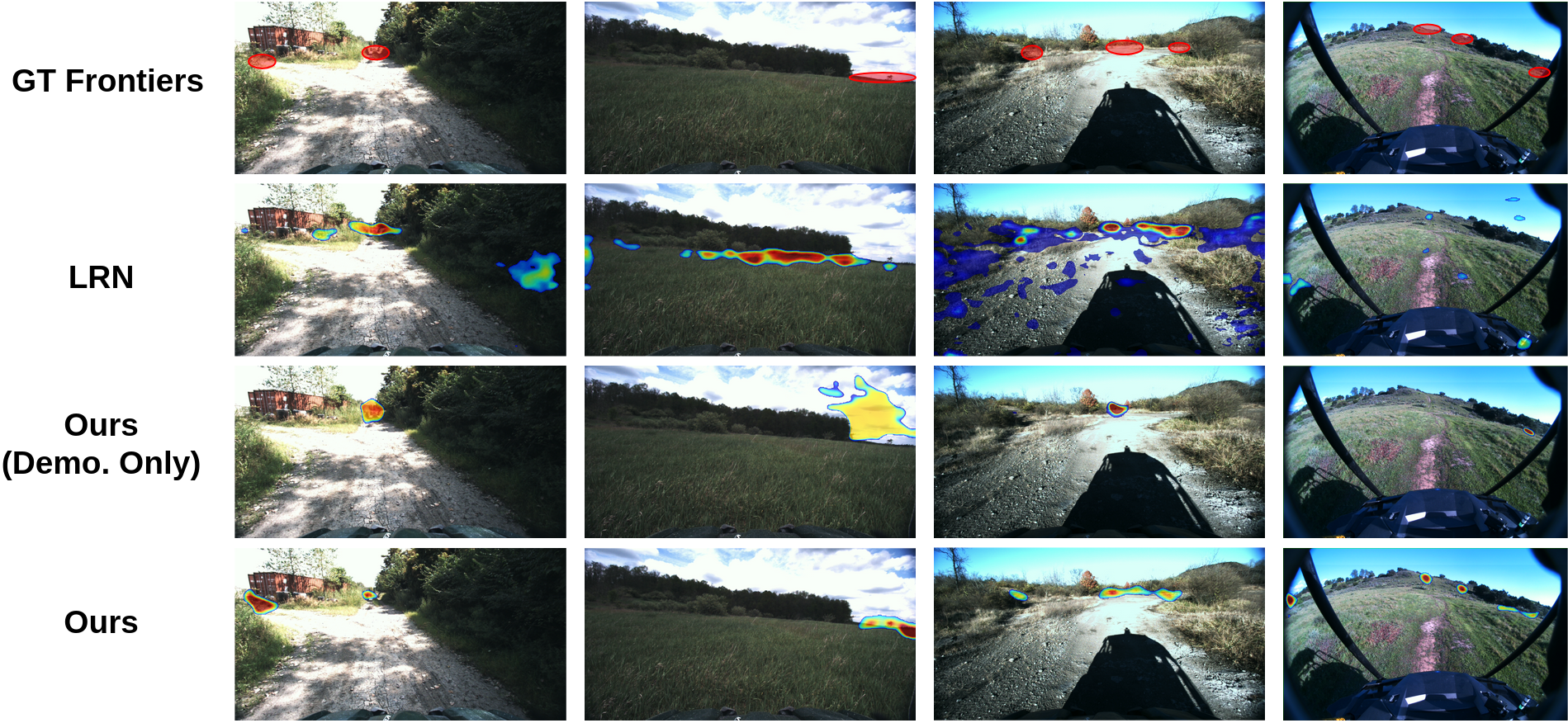}
	\caption{
       Qualitative comparison of our method to LRN. Noisy point tracks and terrain occlusions cause inaccurate training signals for LRN, resulting in false positives (left two columns) and missed frontiers (rightmost column). Similarly, training only using demonstration data causes our method to fail in scenarios with multiple frontiers (middle row). By incorporating planning-based supervision, our method is able to better distinguish multiple viable frontiers.
     }
	\label{fig:img_preds}
    \vspace{-.5cm}
\end{figure*}

Alongside qualitative results, we aim to demonstrate the effectiveness of our approach and use our offline and real-world evaluations to answer the following four questions:
\begin{enumerate}
    \item Does training in the heading space still lead to semantically meaningful predictions in the image space?
    \item Does supplementing human-derived demonstration data with planner-derived demonstration data improve frontier estimation?
    \item Does our method lead to better navigation performance in long-range navigation scenarios?
    \item Does our method generalize to new environments?
\end{enumerate}

\section{Results}

\subsection{Semantically Meaningful Predictions}
We evaluate the semantic meaning of the intermediate image predictions based on performance in the image space classification subtask. Qualitatively, the affordance heatmaps produced by our method are less noisy (Fig. \ref{fig:img_preds}). Our DOS model is usually able to detect one frontier, but the model trained using planning-based labels is more effective in scenarios with multiple frontiers. Our quantitative metrics reflect this as well, as shown in Table \ref{tab:heatmap_metrics}. This is notable considering the model is not supervised in the image space, and tells us that it is still able to learn which parts of an image are most-relevant for long-range navigation. 
While the metrics are lower than usual for segmentation tasks, our method still performs best overall. We primarily use this metric to demonstrate interpretability and potential ease of deployment on different camera systems (demonstrated in latter sections), and use our real-world experiments to evaluate success on the downstream navigation tasks it was optimized for.




\begin{table}[t]
\centering
\caption{Image-space Affordance Classification Metrics}
\begin{tabular}{llccc}
\toprule
Environment & Metric & LRN & Ours Demo. Only & Ours \\
\midrule
\multirow{5}{*}{Similar}
 & AUROC $\uparrow$ & 0.61 & 0.46 & \textbf{0.69} \\
 & AUPRC $\uparrow$ & 0.12 & 0.15 & \textbf{0.25} \\
 & F1 $\uparrow$    & 0.18 & 0.27 & \textbf{0.35} \\
 & Precision\ $\uparrow$& 0.14 & 0.31 & \textbf{0.43} \\
 & Recall\ $\uparrow$ & \textbf{0.39} & 0.36 & 0.35 \\
\midrule
\multirow{5}{*}{OOD}
 & AUROC $\uparrow$ & 0.56 & 0.35 & \textbf{0.61} \\
 & AUPRC $\uparrow$ & 0.06 & 0.04 & \textbf{0.14} \\
 & F1 $\uparrow$    & 0.10 & 0.07 & \textbf{0.24} \\
 & Precision\ $\uparrow$& 0.08 & 0.11 & \textbf{0.21} \\
 & Recall\ $\uparrow$ & 0.17 & 0.07 & \textbf{0.24} \\
\bottomrule
\end{tabular}

\label{tab:heatmap_metrics}
\vspace{-.4cm}
\end{table}


\subsection{Impact of Augmenting Demonstration Data}
To evaluate the impact of training with planner-generated affordances in addition to demonstration data, we compare the data efficiency and real-world performance of a model trained using our method to models trained only with demonstration data: LRN for image-space supervision and DOS for heading-space supervision for comparison. As shown in Tables \ref{tab:heatmap_metrics} and \ref{tab:offline_metrics} the models trained on demonstration data are not as effective at detecting frontiers, especially in unseen environments.

\begin{figure*}[]
	\centering
	\includegraphics[width=.9\linewidth]{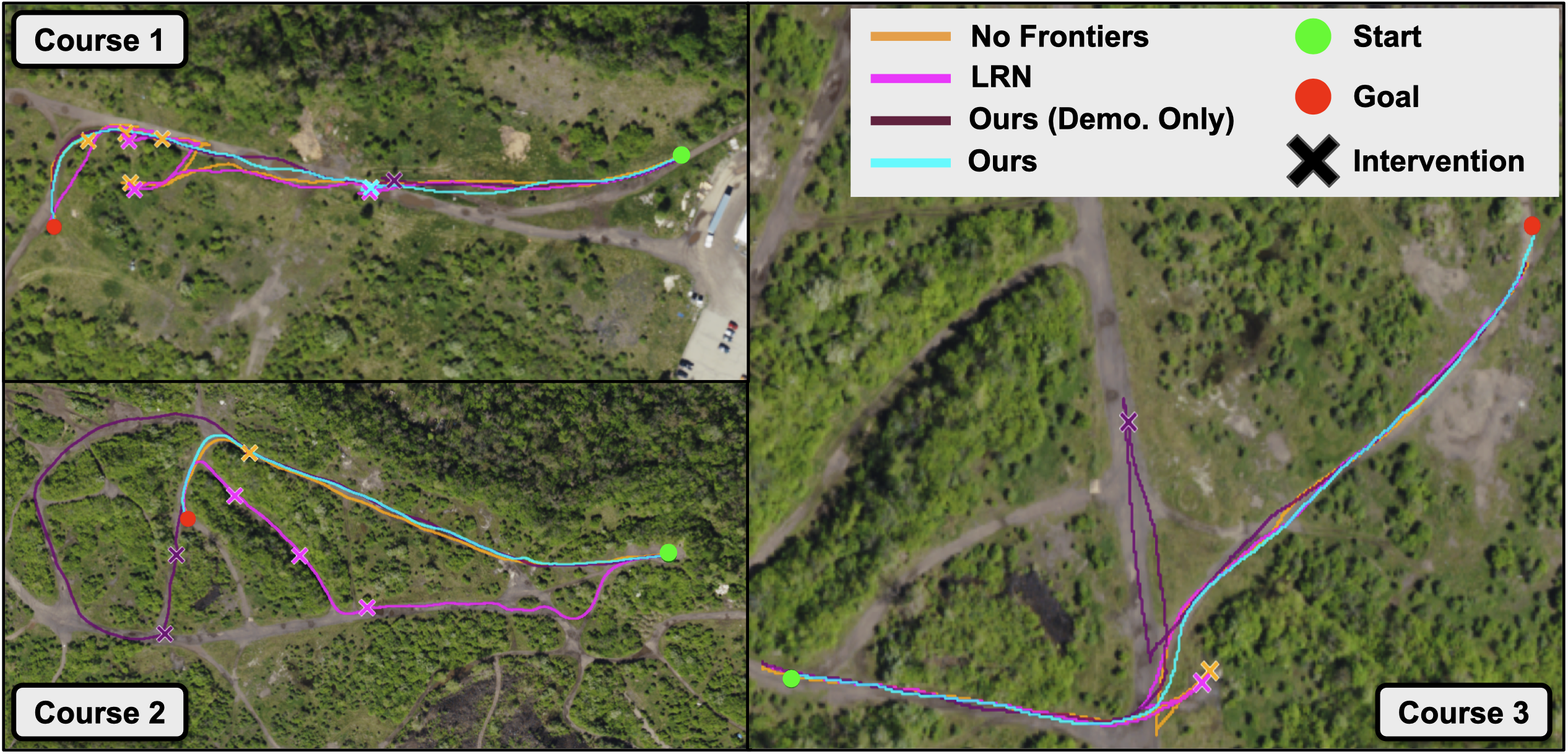}
	\caption{
       Resulting paths taken by our method and baselines in a subset of our real-world experiments. Our method incurs the fewest total interventions and takes the least total time to complete the courses.
     }
	\label{fig:real_world_result}
    \vspace{-.5cm}
\end{figure*}

We also observe better performance in the real-world experiments. Our method resulted in less overall interventions when trained with the planner-generated affordances (Table \ref{tab:rw_metrics}). Moreover, the demonstration-only model was found to be too conservative, as in some scenarios it failed to detect the frontiers that led to the shortest path to the goal (Courses 2 and 3 in Fig \ref{fig:real_world_result}).

\begin{table}[t]
\centering
\caption{Offline Frontier Estimation Metrics}
\begin{tabular}{llcccc}
\toprule
Metric Type & Metric & LRN & DOS & Ours & NFE\\
\midrule
\multirow{3}{*}{Classification}
 & F1 $\uparrow$    & 0.13 & 0.12 & \textbf{0.15} & --\\
 & Precision\ $\uparrow$& 0.09 & \textbf{0.11} & 0.09 & -- \\
 & Recall\ $\uparrow$ & .47 & 0.26 & \textbf{0.76} & --\\
\midrule
\multirow{2}{*}{Goal-Conditioned}
 & Heading Error $\downarrow$ & 0.28 & 0.56 & \textbf{0.22} & 0.40\\
 & MHD $\downarrow$ & 4.04 & 8.9 & \textbf{3.76} & 5.5\\
 
\bottomrule
\end{tabular}
\label{tab:offline_metrics}
\vspace{-.6cm}
\end{table}



\subsection{Real-world Performance}

In our set of five real-world experiments, we find that our method outperforms all baselines in terms of number of interventions and mission completion time. While LRN also outperforms NFE, we find that the predictions are both noisier and overconfident in some scenarios, causing it to get stuck in cul-de-sacs (Courses 1 and 3 in Fig. \ref{fig:real_world_result}). The interventions incurred by our model trained only on demonstration data were due to underconfidence, where it failed to score some frontiers highly enough. While it didn't get stuck in cul-de-sacs, this shortcoming prevented it from exiting main trails when necessary. 

\begin{table}[t]
\centering
\caption{Total Number of Interventions and Completion Time Across All Real-world Experiments}
\begin{tabular}{llccc}
\toprule
Method & Interventions & Time (s) \\
\midrule
 No Frontiers  & 16 & 504.38 \\
 LRN     & 8 & 505.75\\
 Ours Demo. Only & 6 & 519.15 \\
 Ours & \textbf{2} & \textbf{325.84} \\
\bottomrule
\end{tabular}
\label{tab:rw_metrics}
\vspace{-.6cm}
\end{table}

\subsection{Generalizability}
We aim to evaluate the effectiveness of our model in unseen scenarios. While we were unable to perform in-the-loop hardware experiments on a different platform we were able to collect data with a Polaris RZR Turbo S4 ATV, which has a camera configuration and off-road capability that are much different to the platform our model was trained on (example imagery in Fig. \ref{fig:img_preds}). We perform the same set of image-space offline evaluations as before using this data (OOD section of Table \ref{tab:heatmap_metrics}), in addition to our heading space metrics (\ref{sec:heading_space_tasks}) as shown in Table \ref{tab:offline_metrics}. We find that our training framework results in a model that is able to detect frontiers more accurately than the baselines despite being presented with an environment with significantly different topography and self-obstructions.




\section{Conclusions and Future Work}
We present a framework for learning image-based affordances by supervising with global traversability information. The two key decisions core to our approach are directly supervising the heading-space representation that is used for downstream planning, and planning-based data augmentation to supplement our demonstration data. The planner for this process is enabled by first computing vehicle-specific traversability maps from satellite imagery. Training a model in this manner results in more accurate frontier estimation and better performance in real-world off-road navigation experiments, while still producing interpretable intermediate outputs in the image space.

This approach is not without limitations. Specifically, it is important to highlight the reliance on accurate prior maps. If the terrain in the satellite imagery differs greatly from what the robot experiences, the supervision signal is less likely to be accurate. For example, our test environment differed from our map used for training in terms of overhead canopies, erosion, and seasonal changes. Our experiments show that the model can be robust to these discrepancies, but intuitively its performance should improve further with cleaner supervision. 

We believe aggregating model predictions over time into a memory mechanism is an important avenue of future work, as once a frontier is out of sight the system currently has no way to remember that it exists. This could be accomplished either by building a graph-based representation based on predicted frontier headings, or using a map representation such as RayFronts \cite{alama2025rayfronts} to directly map the affordance heatmaps themselves.









\balance

\bibliographystyle{IEEEtran}
\bibliography{IEEEabrv, refs}

\end{document}